\documentclass{article}

\PassOptionsToPackage{numbers}{natbib}

\usepackage[preprint]{neurips_2023}

\usepackage[utf8]{inputenc} 
\usepackage[T1]{fontenc}    
\usepackage{hyperref}       
\usepackage{url}            
\usepackage{booktabs}       
\usepackage{amsfonts}       
\usepackage{nicefrac}       
\usepackage{microtype}      
\usepackage{xcolor}         
\usepackage{xspace}
\usepackage{graphicx}

\usepackage{algorithmic}
\usepackage{algorithm}
\usepackage{Definitions}
\usepackage{wrapfig}

\hypersetup{colorlinks=true,linkcolor=red,citecolor=brown,urlcolor=blue}

\usepackage{color,xcolor}
\usepackage{epsfig}
\usepackage{graphicx}

\usepackage{adjustbox}
\usepackage{array}
\usepackage{booktabs}
\usepackage{colortbl}
\usepackage{float,wrapfig}
\usepackage{hhline}
\usepackage{multirow}
\usepackage{subcaption} 
\usepackage[font={small}]{caption}
\usepackage{natbib}

\usepackage{bm}
\usepackage{nicefrac}
\usepackage{microtype}
\usepackage{mathtools}

\usepackage{changepage}
\usepackage{extramarks}
\usepackage{fancyhdr}
\usepackage{lastpage}
\usepackage{setspace}
\usepackage{xspace}

\usepackage{url}

\usepackage{algorithmic}
\usepackage{algorithm}
\usepackage{todonotes} 
\usepackage{enumitem}
\usepackage{titlesec}
\usepackage{bbm}

\newcommand{\sect}[1]{Section~\ref{#1}}

\newcommand{\eqn}[1]{Equation~\ref{#1}}
\newcommand{\fig}[1]{Figure~\ref{#1}}
\newcommand{\tbl}[1]{Table~\ref{#1}}

\newcommand{\myparagraph}[1]{\vspace{-8pt}\paragraph{#1}}

\newcommand{\model}{Video Adapter\xspace}

\def\methodname{Video Adapter\xspace}

\title{Probabilistic Adaptation of Text-to-Video Models}

\author{%
  Mengjiao Yang\thanks{ Equal contribution. Correspondence to: Mengjiao Yang <sherryy@berkeley.edu>, Yilun Du <yilundu@mit.edu>.}$^{*,1,2}$, 
  \textbf{\,Yilun Du}$^{*,1,3}$, \textbf{Bo Dai}$^{1}$, \\
   \textbf{Dale Schuurmans}$^{1,4}$, \textbf{Joshua B. Tenenbaum}$^{3}$, \textbf{Pieter Abbeel}$^{2}$\\
  $^{1}$Google DeepMind, $^{2}$UC Berkeley, $^{3}$MIT, $^{4}$University of Alberta\\
  \tt\href{https://video-adapter.github.io/}{video-adapter.github.io}
}

\begin{document}

\maketitle

\begin{abstract}
    Large text-to-video models trained on internet-scale data have demonstrated exceptional capabilities in generating high-fidelity videos from arbitrary textual descriptions. However, adapting these models to tasks with limited domain-specific data, such as animation or robotics videos, poses a significant computational challenge, since finetuning a pretrained large model can be prohibitively expensive. Inspired by how a small modifiable component (e.g., prompts, prefix-tuning) can adapt a large language model to perform new tasks without requiring access to the model weights, we investigate how to adapt a large pretrained text-to-video model to a variety of downstream domains and tasks without finetuning. In answering this question, we propose \emph{\methodname}, which leverages the score function of a large pretrained video diffusion model as a probabilistic prior to guide the generation of a task-specific small video model. Our experiments show that \methodname is capable of incorporating the broad knowledge and preserving the high fidelity of a large pretrained video model in a task-specific small video model using as few as 1.25\% parameters of the pretrained model. \methodname is able to generate high-quality yet specialized videos on a variety of tasks such as animation, egocentric modeling, and modeling of simulated and real-world robotics data. 
    More videos can be found on the website 
    \href{https://video-adapter.github.io/}{https://video-adapter.github.io}.
\end{abstract}


\section{Introduction}\label{sec:intro}

Large text-to-video models with billions of parameters trained on internet-scale data have become capable of generating highly realistic videos from 
general text descriptions~\citep{ho2022imagen,hong2022cogvideo,singer2022make}. When such large text-to-video models are used in specialized domains --- or for specialized tasks such as generating videos of robotic plans~\citep{du2302learning}, animation~\citep{wang2019learning}, or videos with customized styles similar to those common in text-to-image~\citep{wang2019learning,ramesh2022hierarchical,liu2022compositional,gal2022image,ruiz2022dreambooth,zhang2023adding} --- a pretrained text-to-video model often requires task-specific adaptation. 
Enabling efficient and effective adaptation of a pretrained text-to-video model to a task-specific setting is one of the major bottlenecks in applying text-to-video in real-world problems.

\begin{figure}[t]
\includegraphics[width=\textwidth]{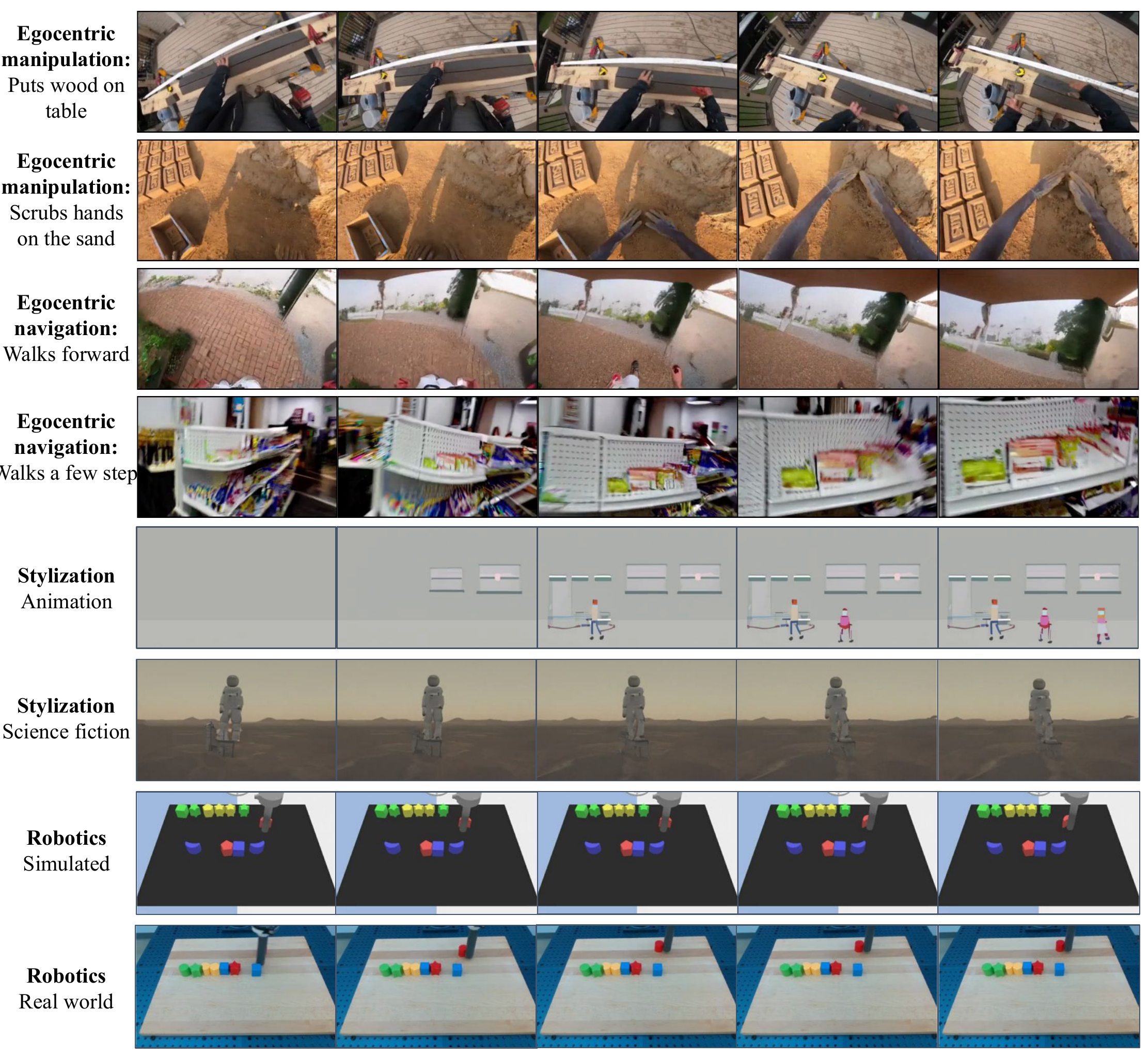}
\caption{\textbf{\methodname Generated Videos.} \methodname is capable of flexible generation of diverse videos with distinct styles including videos with manipulation and navigation based egocentric motions, videos with personalized styles such as animation and science fictions, and simulated and real robotic videos.}
\label{fig:demo}
\end{figure}

Similar to state-of-the-art language models \citep{openai2023gpt4,chowdhery2022palm}, the parameters of a pretrained text-to-video model are usually not publicly available \citep{ho2022imagen,villegas2022phenaki,blattmann2023align}, which makes finetuning on task-specific data infeasible. Even if finetuning access to these large models can be gained through API calls to an industrial service~\citep{openai2023gpt4}, the computational resources involved in finetuning can be prohibitively expensive, hindering the research progress of small industrial and academic institutions. While some techniques for controllable text-to-image generation have been developed --- such as light-weight training methods like ControlNet~\citep{zhang2023adding} and training-free methods such as manipulating CLIP features~\citep{ramesh2022hierarchical}, or using a small set of images with same topics or objects~\citep{gal2022image,ruiz2022dreambooth} --- it is not immediately clear whether the techniques developed for text-to-image can directly transfer to text-to-video, especially given that generating videos is much more complex than generating images~\citep{li2018video} and that pretrained text-to-video models are often orders of magnitude larger than text-to-image models~\citep{blattmann2023align}.

Inspired by the progression of task-specific adaptation in language modeling from finetuning to few shot in-context learning~\citep{brown2020language} then to more sophisticated prompting schemes~\citep{wei2022chain} as pretrained language models became more capable and expensive, we ask the natural question of whether it is possible to adapt a large pretrained text-to-video model using a small controllable component similar to prefix tuning of language models~\citep{li2021prefix} and thus avoid expensive finetuning of large pretrained models. Intuitively, even though the video statistics in a downstream task (e.g., robotics, animation) might differ significantly from the pretraining videos, certain video properties such as the semantics and dynamics of objects should remain the same between pretraining and adaptation. As a result, a large pretrained video model could be used as a knowledge prior to guide the generation of task-specific videos while maintaining broad properties such as temporal consistency and object permanence.

To this end, we propose \methodname, a probabilistic approach for exploiting a large pretrained video diffusion model to guide the generation of task or domain specific videos. 
By factoring the domain-specific video distribution into a pretrained prior and a small trainable component, the cost of adapting a large pretrained video model can be significantly reduced, while enabling generation of videos that satisfy the characteristics of both the pretrained prior (i.e., temporal consistency and object permanence) and the styles or properties unique to the videos of interest (e.g., animation).
We evaluate \methodname on a diverse set of video generation tasks as illustrated in Figure~\ref{fig:demo}. 
Quantitatively, \methodname generates videos that obtain better FVD and Inception Scores than a high quality pretrained large video model while using up to 80x fewer parameters on the robotic Bridge data~\citep{ebert2021bridge} and the challenging Ego4D data~\citep{grauman2022ego4d}. 
Qualitatively, we show that \methodname enables generation of stylized videos such as sci-fi and animation. 
Furthermore, we demonstrate how \methodname can pave the way for bridging the notorious sim-to-real gap in robotics~\citep{zhao2020sim} by modeling both real and simulated robotic videos while enabling data augmentation on real robotic videos through customized stylisation.

\section{Preliminaries}
\label{sect:diffusion}

We first introduce relevant background information on denoising diffusion probabilistic models (DDPMs) and discuss their connection to Energy-Based Models (EBMs). We will then use this connection to EBMs to convert large text-to-video diffusion models to probabilistic priors.

\paragraph{Denoising Diffusion Probabilistic Models.}
Denoising diffusion probablistic models~\citep{sohl2015deep,ho2020denoising} are a class of probabilistic generative models where the generation of a video $\tau = [x_1, \ldots, x_H] \in X^{H}$ is formed by iterative denoising. Given a video $\tau$ sampled from a video distribution $p(\tau)$, a randomly sampled Gaussian noise variable $\epsilon \sim \mathcal{N}(\textbf{0}, \textbf{I})$,   
and a set of $T$ different noise levels $\beta_t$, a denoising model $\epsilon_\theta$ is trained to denoise the noise corrupted video $\tau$ at each specified noise level $t \in [1, T]$:
\begin{equation*}
    \label{eq:diffusion_loss}
    \mathcal{L}_{\text{MSE}}=\|\mathbf{\epsilon} - \epsilon_\theta(\sqrt{1 - \beta_t} \tau +  \sqrt{\beta_t} \mathbf{\epsilon}, t))\|^2
\end{equation*}

Given this learned denoising function, new videos may be generated from the diffusion model by initializing a video sample $\tau_T$ at noise level $T$ from a Gaussian $\mathcal{N}(\textbf{0}, \textbf{I})$. This sample $\tau_T$ is then iteratively denoised following the expression:
%
\begin{equation}
    \tau^{t-1}=\alpha^t(\tau^{t}-\gamma^t \epsilon_\theta(\tau^t,t) +\xi),\quad \xi\sim \mathcal{N} \bigl(\textbf{0}, \sigma^2_t \textbf{I} \bigl),
    \label{eq:unconditional_langevin}
\end{equation}
where $\gamma^t$ is the step size of denoising and $\alpha^t$ is a linear decay on the currently denoised sample. The final sample $\tau_0$ after $T$ rounds of denoising corresponds to the final generated video. 

\paragraph{Energy-Based Models View of DDPMs.}
The denoising function $\epsilon_\theta$ estimates the score~\cite{vincent2011connection, song2019generative, liu2022compositional} of an underlying (unnormalized) EBM probability distribution~\cite{lecun2006tutorial,du2019implicit} characterizing the noise perturbed data.  Therefore, a diffusion model corresponds to an EBM, $p_\theta(\tau) \propto e^{-E_\theta(\tau)}$, where the denoising function is given by $\epsilon(\tau^t,t) = \nabla_{\tau} E_{\theta}(\tau^{t})$. The sampling procedure in a diffusion model corresponds to the Langevin sampling procedure on an EBM (see derivation in Appendix~\ref{app:diffusion-ebm}):
\begin{equation}
    \tau^{t-1}= \alpha^t(\tau^{t}-\gamma \nabla_{\tau} E_{\theta}(\tau^{t}) + \xi),\quad \xi\sim \mathcal{N} \bigl(\textbf{0}, \sigma^2_t \textbf{I} \bigl).
    \label{eq:unconditional_ebm}
\end{equation}
This equivalence of diffusion models and EBMs allows us to consider sampling from the product of two different diffusion models $p_1(\tau)p_2(\tau)$, such that each diffusion model corresponds to an EBM, $e^{-E_1(\tau)}$ and $e^{-E_2(\tau)}$, and the product is given by $e^{-E'(\tau)} = e^{-(E_1(\tau) + E_2(\tau))}$. In particular, we can sample from this new distribution also by using Langevin sampling:
\begin{equation}
    \tau^{t-1}= \alpha^t(\tau^{t}-\gamma \nabla_{\tau} E'_{\theta}(\tau^{t}) + \xi),\quad \xi\sim \mathcal{N} \bigl(\textbf{0}, \sigma^2_t \textbf{I} \bigl),
    \label{eq:compose_ebm}
\end{equation}
which corresponds to the sampling procedure using denoising functions
\begin{equation}
    \tau^{t-1}= \alpha^t(\tau^{t}-\gamma (\epsilon_\theta^1(\tau^t,t) + \epsilon_\theta^2(\tau^t, t)) + \xi),\quad \xi\sim \mathcal{N} \bigl(\textbf{0}, \sigma^2_t \textbf{I} \bigl).
    \label{eq:compose_ebm_realize}
\end{equation}
Below we will illustrate how this factored EBM parameterization of a diffusion model can allow a large pretrained text-to-video model to be leveraged as a probabilistic prior.





\section{Probabilistic Adaptation of Text-to-Video Models}\label{sec:method}
\begin{figure}[t]
\includegraphics[width=\textwidth]{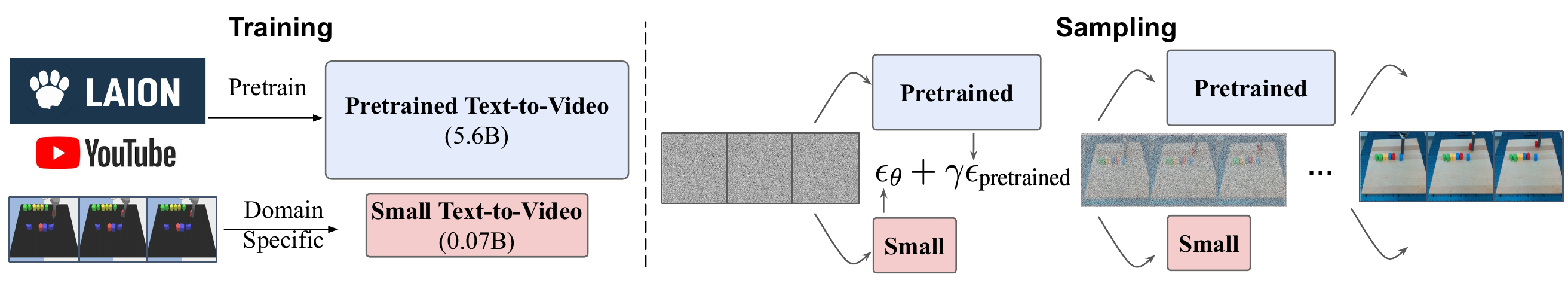}
\caption{\textbf{\methodname Framework.} \methodname only requires training a small domain-specific text-to-video model with orders of magnitude fewer parameters than a large video model pretrained from internet data. During sampling, \methodname composes the scores of the pretrained and the domain specific video models, achieving high-quality and flexible video synthesis.}
\label{fig:framework}
\end{figure}



To explain how a large text-conditioned video diffusion model can be effectively used as a probabilistic prior for video generation,
we will first introduce the functional form we consider in \sect{sect:prior},
then discuss how the probabilistic composition can be implemented with diffusion models in \sect{sect:prob_adapt}.
To generate high-quality videos, we also explain in \sect{sect:low_temperature} how the underlying probabilistic composition can be sharpened generate low temperature samples.

\subsection{Large Text-to-Video Models as Probabilistic Priors}
\label{sect:prior}

Large text-to-video models are now pretrained on massive datasets consisting of millions of videos, and are therefore able to capture a powerful prior 
$p_{\text{pretrained}}(\tau|\text{text})$ on the natural distribution of videos $\tau$. 
Such a distribution $p_{\text{pretrained}}(\tau|\text{text})$ encodes ubiquitous characteristics that are shared across videos such as temporal consistency, object permanence, and the underlying semantics of different objects.

Given a smaller new dataset of video-text pairs, $D_{\text{Adapt}} = \{ (\tau_0, \text{text}_0), (\tau_1, \text{text}_1), \ldots, (\tau_T, \text{text}_T) \}$,
how can one leverage the powerful prior captured by a pretrained video diffusion model to synthesize videos 
similar to those in
$D_{\text{Adapt}}$? 
One approach is to directly finetune the weights of 
$p_{\text{pretrained}}(\tau|\text{text})$
on the underlying videos of $D_{\text{Adapt}}$. 
Such a direct approach however has several drawbacks: \textbf{(1)} The underlying pretrained diffusion model will often have billions of parameters, making it difficult to even the load model weights in a consumer grade GPU, let alone finetuning. \textbf{(2)} The underlying weights of the network may further not be publicly accessible given different proprietary and industrial constraints.


Similar challenges with large language models have led to the 
creation of training-free methods to adapt language models to new datasets $D_{\text{Adapt}}$, where different prompts and in-context demonstrations are used to illicit desired language generation in language models. 
Analogously, we propose \methodname as a training-free method to adapt pretrained video diffusion to a new dataset of videos $D_{\text{Adapt}}$ through probabilistic composition. 
Given $D_{\text{Adapt}}$, we learn a separate small video diffusion model $p_{\theta}(\tau|\text{text})$ to represent the distribution of videos in $D_{\text{Adapt}}$. 
We then adapt $p_{\text{pretrained}}(\tau|\text{text})$ to  $D_{\text{Adapt}}$ by constructing a product distribution $p_{\text{joint}}(\tau|\text{text})$ in the form:
\begin{equation}
    \label{eqn:probabilistic_prior}
    \underbrace{p_{\text{product}}(\tau|\text{text})}_{\text{Product Distribution}} \propto \hskip.5em \underbrace{p_{\text{pretrained}}(\tau|\text{text})}_{\text{Pretrained Prior}} \hskip.25em \underbrace{p_{\theta}(\tau|\text{text})}_{\text{Video Model}}.
\end{equation}
By fixing the pretrained model $p_{\text{pretrained}}(\tau|\text{text})$, we train the video model $p_{\theta}(\tau|\text{text})$ via maximum likelihood estimation on $D_{\text{Adapt}}$. 
This allows $p_{\theta}(\tau|\text{text})$ to exhibit high likelihood across videos in $D_{\text{Adapt}}$, but because it is a small model trained on less diverse data it can also exhibit erroneously high likelihood across many unrealistic videos. The product distribution $p_{\text{product}}(\tau|\text{text})$ removes unrealistic videos by downweighting any videos $\tau$ that not likely under the pretrained prior, enabling one to controllably generate videos in the style in $D_{\text{Adapt}}$.


\subsection{Implementing Probabilistic Adaptation}
\label{sect:prob_adapt}

To adapt the large pretrained text-to-image model $p_{\text{product}}(\tau|\text{text})$ from \eqn{eqn:probabilistic_prior}, as well as to sample from it, we exploit the EBM interpretation of diffusion models discussed in \sect{sect:diffusion}.
Based on the EBM interpretation, the pretrained diffusion model $p_{\text{pretrained}}(\tau|\text{text})$ corresponds to an EBM $e^{-E_{\text{pretrained}}(\tau|\text{text})}$ while the smaller video model $p_{\theta}(\tau|\text{text})$ parameterizes an EBM $e^{-E_{\theta}(\tau|\text{text})}$. The product distribution then corresponds to:
\begin{equation*}
    p_{\text{product}}(\tau|\text{text}) \propto p_{\text{pretrained}}(\tau|\text{text}) p_{\theta}(\tau|\text{text}) \propto e^{-(E_{\text{pretrained}}(\tau|\text{text}) + E_{\theta}(\tau|\text{text}))} = e^{-E'(\tau|\text{text})},
\end{equation*}
which specifies a new EBM $E'(\tau)$ from the sum of energy functions of the component  models.

Substituting this EBM into \eqn{eq:compose_ebm} shows that one can sample from the product distribution $p_{\text{product}}(\tau|\text{text})$ through the diffusion sampling procedure:
\begin{equation*}
    \tau^{t-1}= \alpha^t(\tau^{t}-\gamma \nabla_{\tau}  (E_{\text{pretrained}}(\tau^{t}|\text{text}) +E_{\theta}(\tau^{t}|\text{text})) + \xi),\quad \xi\sim \mathcal{N} \bigl(\textbf{0}, \sigma^2_t \textbf{I} \bigl)
\end{equation*}
which corresponds to sampling from \eqn{eq:unconditional_langevin} according to
\begin{equation*}
    \tau^{t-1}= \alpha^t(\tau^{t}-\gamma (\epsilon_{\text{pretrained}}(\tau^t,t|\text{text}) + \epsilon_\theta(\tau^t,t|\text{text})) + \xi),\quad \xi\sim \mathcal{N} \bigl(\textbf{0}, \sigma^2_t \textbf{I} \bigl).
\end{equation*}

Thus, to probabilistically adapt a pretrained text-to-video model to a new dataset $D_{\text{Adapt}}$, we use the standard diffusion sampling procedure, but where the denoising prediction is the sum of predictions from both the pretrained and small video diffusions. To control the strength of the pretrained prior in on our final video generation, we scale the pretrained distribution by an inverse temperature $\gamma$ to obtain $p_{\text{pretrained}}^\gamma(\tau|\text{text})$, which corresponds to scaling the denoised prediction from $\epsilon_{\text{pretrained}}(\tau^t,t|\text{text})$ by a scalar $\gamma$
\begin{equation*}
    \epsilon(\tau^t, t|\text{text}) = \epsilon_\theta(\tau^t, t|\text{text}) + 
    \gamma\epsilon_{\text{pretrained}}(\tau^t, t|\text{text}). 
\end{equation*}

A similar approach to combining the scores of multiple instances of a single diffusion model was used to enable compositional generation in ~\citep{liu2022compositional}. 
Probabilistic adaption from the combined model can be further improved by integrating multiple steps MCMC sampling between each diffusion noise distribution as done in ~\citep{du2023reduce}.


\subsection{Low Temperature Sampling}
\label{sect:low_temperature}

In practice, directly denoising videos from a denoising network $\epsilon(\tau^t, t)$ generates poor videos, as the underlying learned distribution $p_\theta(\tau)$ exhibits too many spurious likelihood modes. To generate sharp videos with diffusion models, low temperature video samples are generated by utilizing classifier free guidance~\citep{ho2022classifier}, which corresponds to sampling from the modified probability composition:
\begin{equation*}
    p^{\text{cfg}}(\tau|\text{text}) \;\propto\; p(\tau) \left (\frac{p(\tau|\text{text})}{p(\tau)} \right)^{\alpha} \;\propto \;p(\tau) p(\text{text}|\tau)^{\alpha},
\end{equation*}
where $\alpha$ corresponds to the classifier free guidance score, typically chosen to be significantly larger than 1. By upweighting the expression $p(\text{text}|\tau)$ via the inverse temperature $\alpha$ the modified distribution $p^{cfg}(\tau|\text{text})$ above generates lower temperature video samples conditioned on the text.  

It appears straightforward to similarly construct low temperature samples from our proposed composition by sampling from the distribution
\begin{equation*}
    p^{\text{cfg}}_{\text{product}}(\tau|\text{text}) \propto p^{\text{cfg}}_{\text{pretrained}}(\tau|\text{text}) p^{\text{cfg}}_{\theta}(\tau|\text{text}),
\end{equation*}
but using the classifier free distribution distribution $p^{\text{cfg}}_{\text{pretrained}}(\tau|\text{text})$ as our probabilistic prior of video is now problematic, as it now has very few high probability modes. 


To effectively leverage a broad probabilistic prior while simultaneously generating low temperature samples, we propose to first construct a new text-conditional distribution video distribution following \sect{sect:prior}:
\begin{equation*}
 p_{\text{product}}(\tau | \text{text}) \propto p_{\text{pretrained}} (\tau | \text{text}) p_\theta(\tau | \text{text}).
\end{equation*}
We can then use the density ratio of this modified text-conditioned distribution with the unconditional video density $p_\theta(\tau)$ learned on $D_{\text{Adapt}}$ to construct a new implicit classifier $p_{\text{product}}(\tau | \text{text})$.

By increasing the inverse temperature $\alpha$ on this implicit classifier,  we can generate low temperature and high quality video samples conditioned on a given text by sampling from the modified distribution
\begin{equation*}
    \tilde{p}_{\theta}^{*}(\tau|\text{text}) = p_{\theta}(\tau) \left (\frac{\tilde{p}_{\theta}(\tau|\text{text})}{p_{\theta}(\tau)} \right)^{\alpha},
\end{equation*}
which corresponds to sampling from a modified denoising function
\begin{equation*}
    \tilde{\epsilon}_{\theta}(\tau, t|\text{text}) = \epsilon_{\theta}(\tau, t) + \alpha (\epsilon_{\theta}(\tau, t|\text{text}) + \gamma \epsilon_{\text{pretrained}}(\tau, t|\text{text}) - \epsilon_{\theta}(\tau, t))
\end{equation*}
We quantitatively and qualitatively ablate the effect of this denoising function in \fig{fig:ablation} and \tbl{tbl:ablations},
showing that this variant leads to better blending of styles between models. 
The overall pseudocode for the proposed approach with classifier-free guidance is given in Algorithm \ref{alg:sample}.

\begin{algorithm}[t]
\small
\begin{algorithmic}
    \STATE \textbf{Input:} Pretrained Text-to-Video Model $\epsilon_\text{pretrained}(\tau, t|\text{text})$, Inverse Temperature $\alpha$, Prior strength $\gamma$.
   \STATE Initialize sample  $\tau_T \sim \mathcal{N}(\bm{0}, \bm{I})$  \\
    \FOR{$t = T, \ldots, 1$}
        \STATE $\tilde{\epsilon}_{\text{text}} \gets \epsilon_\theta(\tau_{t}, t | \text{text}) + \gamma \epsilon_{\text{pretrained}}(\tau_{t}, t | \text{text})$
        \hspace{1.8cm} \small{\color{gray}// compute score using text-conditioned prior} \\
        \STATE $\epsilon \gets \epsilon_\theta(\tau_{t}, t)$
        \hspace{7.2cm} \small{\color{gray}// compute unconditional score} \\
        \STATE $\tilde{\epsilon}_{\text{cfg}} \gets \epsilon + \alpha  (\tilde{\epsilon}_{\text{text}} - \epsilon)$ \hspace{4.8cm} \small{\color{gray}// compute classifier guidance score score}  \\
        \STATE $\tau_{t-1} = \texttt{ddpm\_sample}(\tau_t, \tilde{\epsilon}_{\text{cfg}}) $ \hspace{2.8cm} \small{\color{gray}// run diffusion sampling (can use other samplers)} \\
    \ENDFOR \\

  \end{algorithmic}
 \caption{Sampling algorithm of \model}
 \label{alg:sample}
 \end{algorithm}

\section{Experiments}\label{sec:exp}

In this section, we illustrate how a large pretrained text-to-video model 
can deliver a rich set of downstream capabilities
when combined 
with a task-specific video model. 
In particular, leveraging a high quality and broad probabilistic prior enables (1) controllable video synthesis from edge-only inputs, (2) high-quality video modeling that outperforms both the pretrained model and the task-specific video model, and (3) data augmentation between simulated and real robotic videos. See experiment details and additional experimental results in Appendix~\ref{app:exp} and in supplementary material.\footnote{See video visualizations at 
\href{https://video-adapter.github.io/}{https://video-adapter.github.io/.}}

\subsection{Controllable Video Synthesis}

\myparagraph{Setup.} We first demonstrate that the probabilistic prior in \methodname can used to adapt and modify the styles of videos. We curate two adaptation datasets, one with an ``animation'' style and the other with a ``scifi'' style, where videos containing relevant keywords in their descriptions are grouped together to form $D_\text{Adapt}$. To enable controllable video synthesis, we pretrain a large model to map Sobel edges to videos, and similarly train task-specific small models to map Sobel edges to the stylized videos. The pretrained model has 5.6B parameters, whereas the task-specific small models have only 6\% of the pretrained model's parameters (330M).

\begin{figure}[t]
\includegraphics[width=\linewidth]{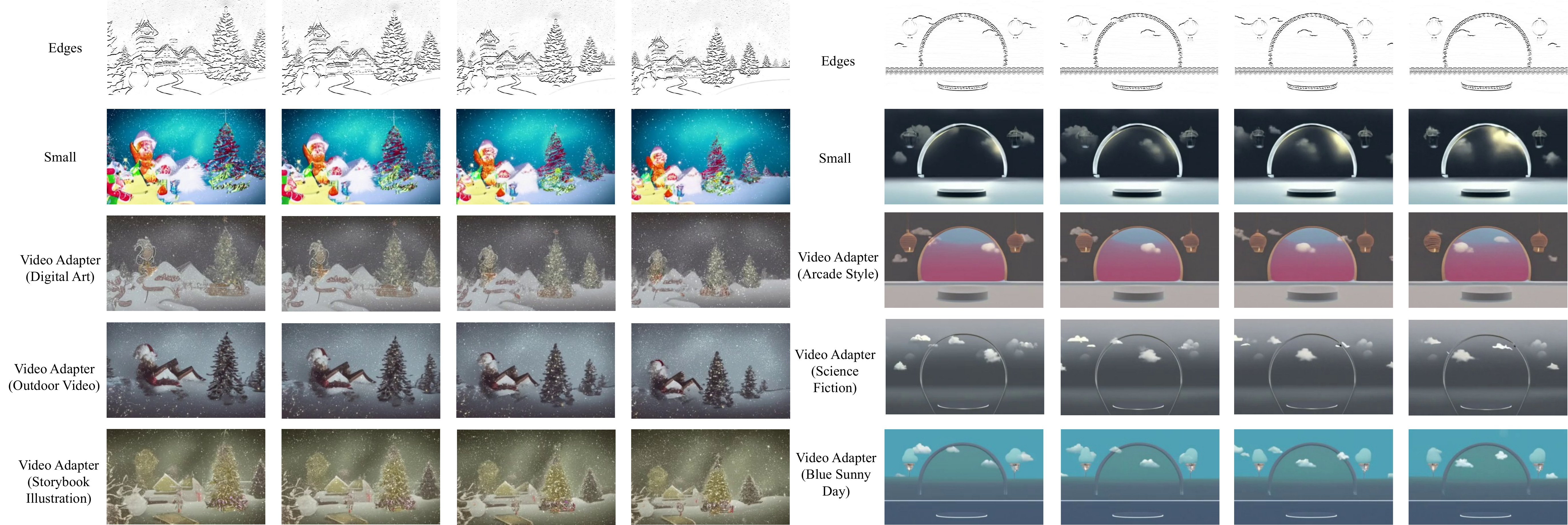}
\caption{\textbf{\methodname enables stylization of a Animation Specific Model.} \methodname enables a large pretrained model to adapt and change the style a small animation style model.}
\label{fig:video_anime}
\end{figure}

\begin{figure}[t]
\includegraphics[width=\linewidth]{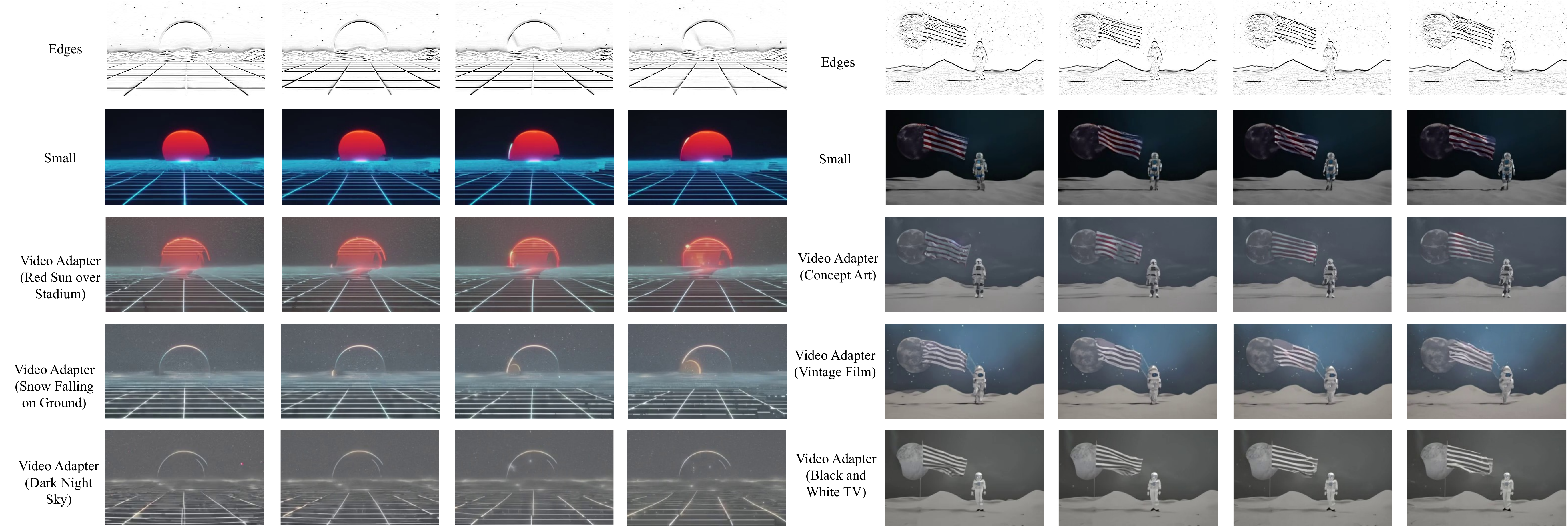}
\caption{\textbf{\methodname enables stylization of a SciFi Specific Model.} \methodname enables a large pretrained model to adapt and change the style a small Scifi animation style model.}
\label{fig:video_scifi}
\end{figure}

\myparagraph{Stylizing Video Generation.} In \fig{fig:video_anime}, we demonstrate how the pretrained prior can adapt an animation style model to generate videos in alternative styles, while maintaining an animated quality. In \fig{fig:video_scifi}, we further demonstrate how the pretrained prior can also adapt the scifi style model to generate videos in other styles.

\begin{wrapfigure}{r}{7.0cm}
\vspace{-5pt}
\includegraphics[width=\linewidth]{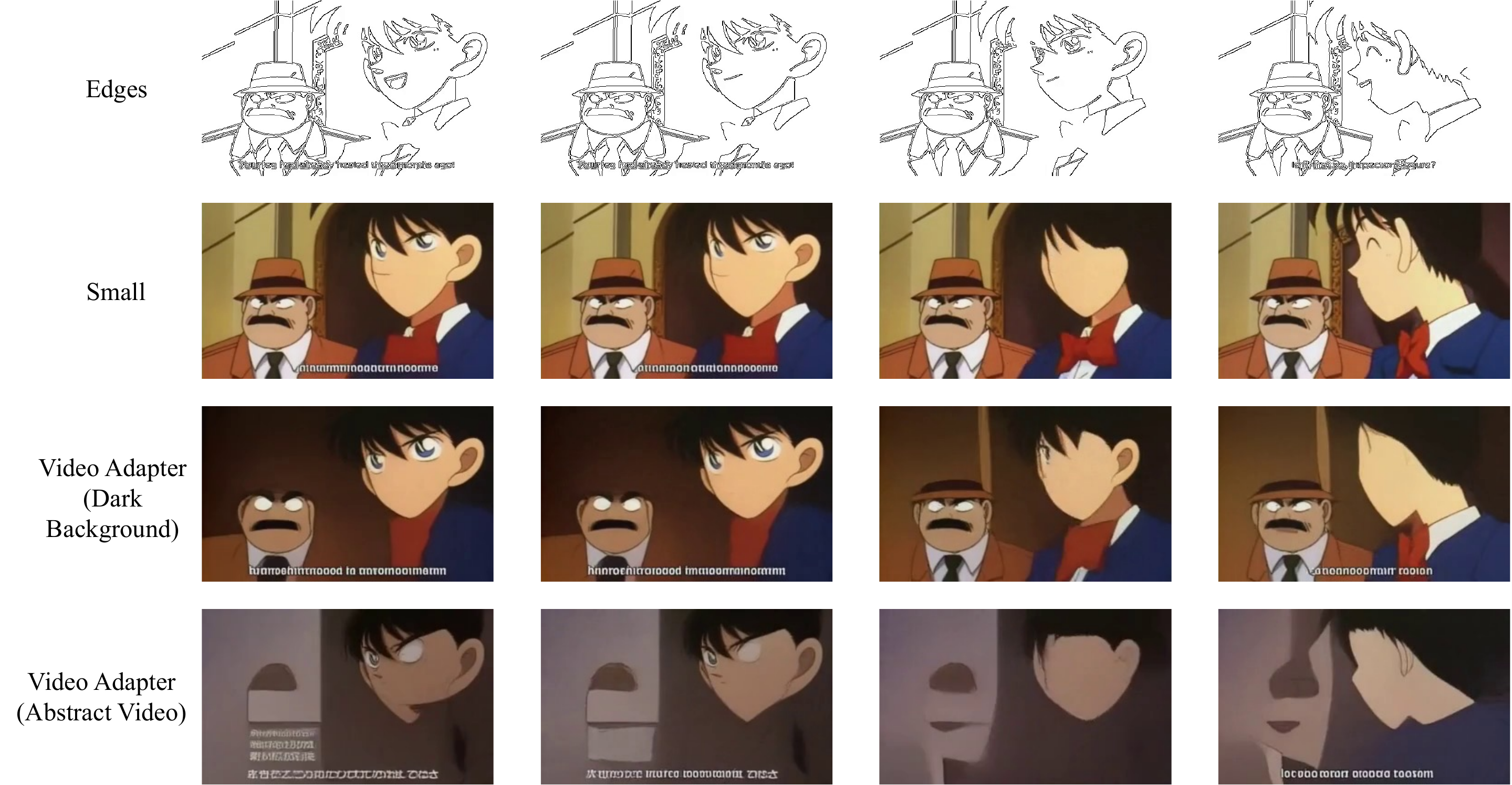}
\caption{\textbf{Instance Specific Stylization.} \methodname enables the stylization of video model trained on a single animation style}
\label{fig:video_animation_specific}
\end{wrapfigure}

\myparagraph{Specific Animation Style.} We further trained a small video model on an ``animation'' style of a particular artist.
In \fig{fig:video_animation_specific}, we illustrate how the pretrained prior model can maintain the artist's style while changing the background.

\begin{figure}[t]
\includegraphics[width=\linewidth]{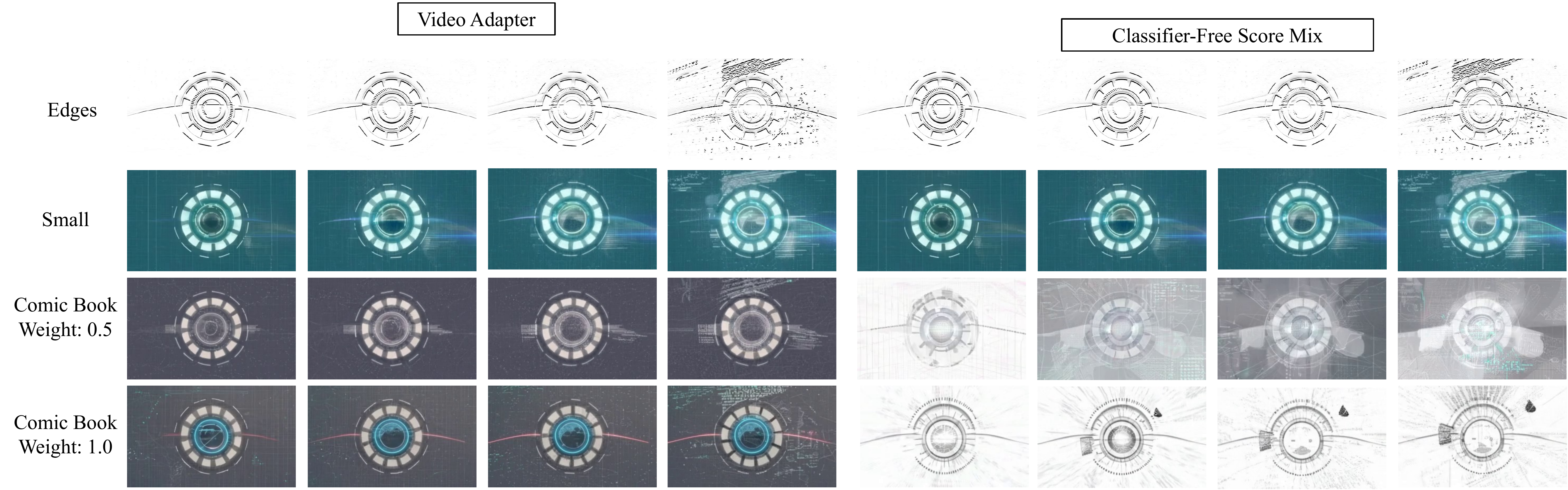}
\caption{\textbf{Analysis of \methodname.} As the adaptation weight on a pretrained prior is increased, \methodname maintains the original style of $D_{\text{Adapt}}$ while also adding style specified by the pretrained model. In contrast, directly interpolating two classifier-free guidance scores leads to incoherent images between styles (weight=0.5) and collapses to the style of the pretrained model (weight=1.0).}
\label{fig:ablation}
\end{figure}

\myparagraph{Analysis.} In \fig{fig:ablation}, we change the magnitude of the weight on the pretrained prior, and compare \methodname with directly interpolating the classifier-free scores between the pretrained and adapter models. We find that \methodname maintains the adapter style more accurately, whereas classifier-free score interpolation collapses to the teacher style with intermediate interpolation, leading to erratic artifacts. 



\subsection{High-Quality Efficient Video Modeling}

\myparagraph{Setup.} To demonstrate \methodname's ability in adapting to video domains that are not a part of pretraining, we consider adapting the pretrained model to two  downstream datasets, Ego4D~\citep{grauman2022ego4d} and Bridge Data~\citep{ebert2021bridge}. These adaptations are nontrivial, as Ego4D consists of mostly egocentric videos that are not commonly found on the internet. Similarly, the Bridge Data consists of task-specific videos of a WidowX250 robot that is out of the distribution of the pretraining data. For Ego4D, we take a subset of the original dataset consisting of 97k text-video pairs and split them into train (90\%) and test (10\%) to form $D_\text{Adapt}$. For the Bridge Data, we take the entire dataset consisting of 7.2k text-video pairs and use the same train-test split to form $D_\text{Adapt}$.

For the pretrained model, we use the 5.6B base model pretrained on generic internet videos from~\citep{ho2022imagen}. For the task-specific small model, we downscale the video diffusion model from~\citep{ho2022imagen} by a factor of 80, 40, and 2 to create a diverse set of small models to be trained on task-specific $D_\text{Adapt}$. Table~\ref{tbl:video} shows the number of parameters of pretrained and small video models. Both the pretrained model and the small models are trained to generate subsequent frames conditioned on the first frame.

\begin{table*}[t]
\small\setlength{\tabcolsep}{5.5pt}
\centering
\begin{tabular}{lcccccc}
      &  \multicolumn{3}{c}{\bf Bridge} & \multicolumn{3}{c}{\bf Ego4D} \\
      \cmidrule(lr){2-4} \cmidrule(lr){5-7} 
      {\bf Model} & FVD $\downarrow$ & FID $\downarrow$ & Param (B)$\downarrow$ & FVD $\downarrow$ & IS $\uparrow$ & Param (B) $\downarrow$ \\
      \midrule
      Small (S) & 186.8 & 38.8 & 0.07 & 228.3 & 2.28 & 0.07 \\
      Small (S) + Pretrained & \textbf{177.4} & \textbf{37.6} & 0.07 & 156.3 & 2.82 & 0.07\\
      Small (L) & 152.5 & 30.1 & 0.14 & 65.1 & 3.31 & 2.8\\
      Small (L) + Pretrained & \textbf{148.1} & \textbf{29.5} & 0.14 & \textbf{52.5} & \textbf{3.53} & 2.8 \\
      Pretrained & 350.1 & 42.6 & 5.6 & 91.7 & 3.12 & 5.6 \\ 
    \bottomrule
\end{tabular}
\caption{\small \textbf{Video Modeling Quantitative Performance} \methodname (Small + Pretrained) achieves better FVD, FID, and Inception Scores than both the pretrained model and the task-specific small model while only training parameters as fewer as 1\% of the pretrained model.}
\label{tbl:video}
\end{table*}

\begin{figure}[t]
\includegraphics[width=\linewidth]{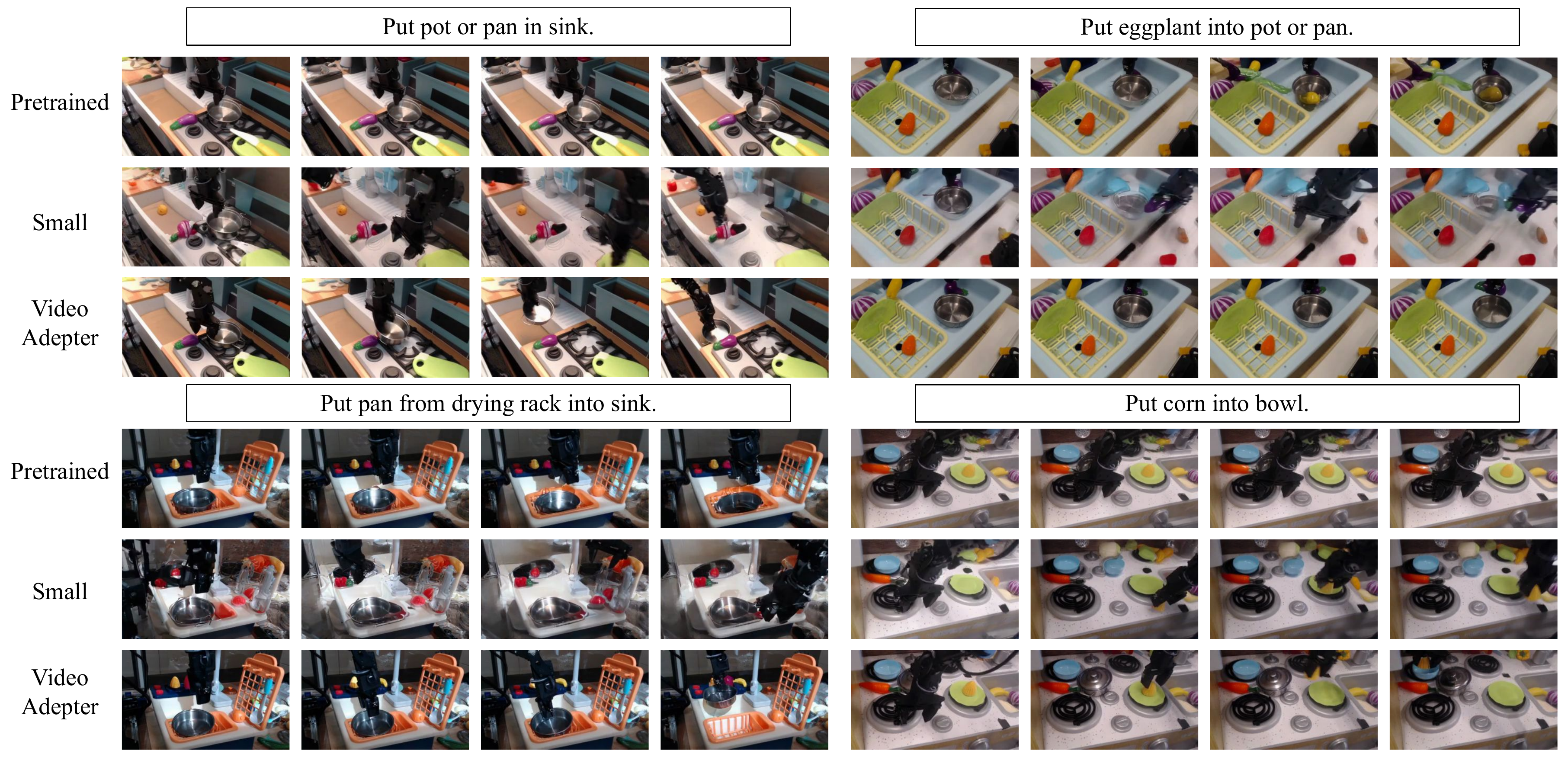}
\caption{\textbf{\methodname on Bridge Data.} The pretrained model (first row) produces videos that are high-quality but are generally static and fail to complete the task. The small model (second row) produces low-quality videos with unrealistic arm movements. \methodname (third row) produces high-quality videos and successfully completes the task.}
\label{fig:video_bridge}
\end{figure}
\begin{figure}[t]
\includegraphics[width=\textwidth]{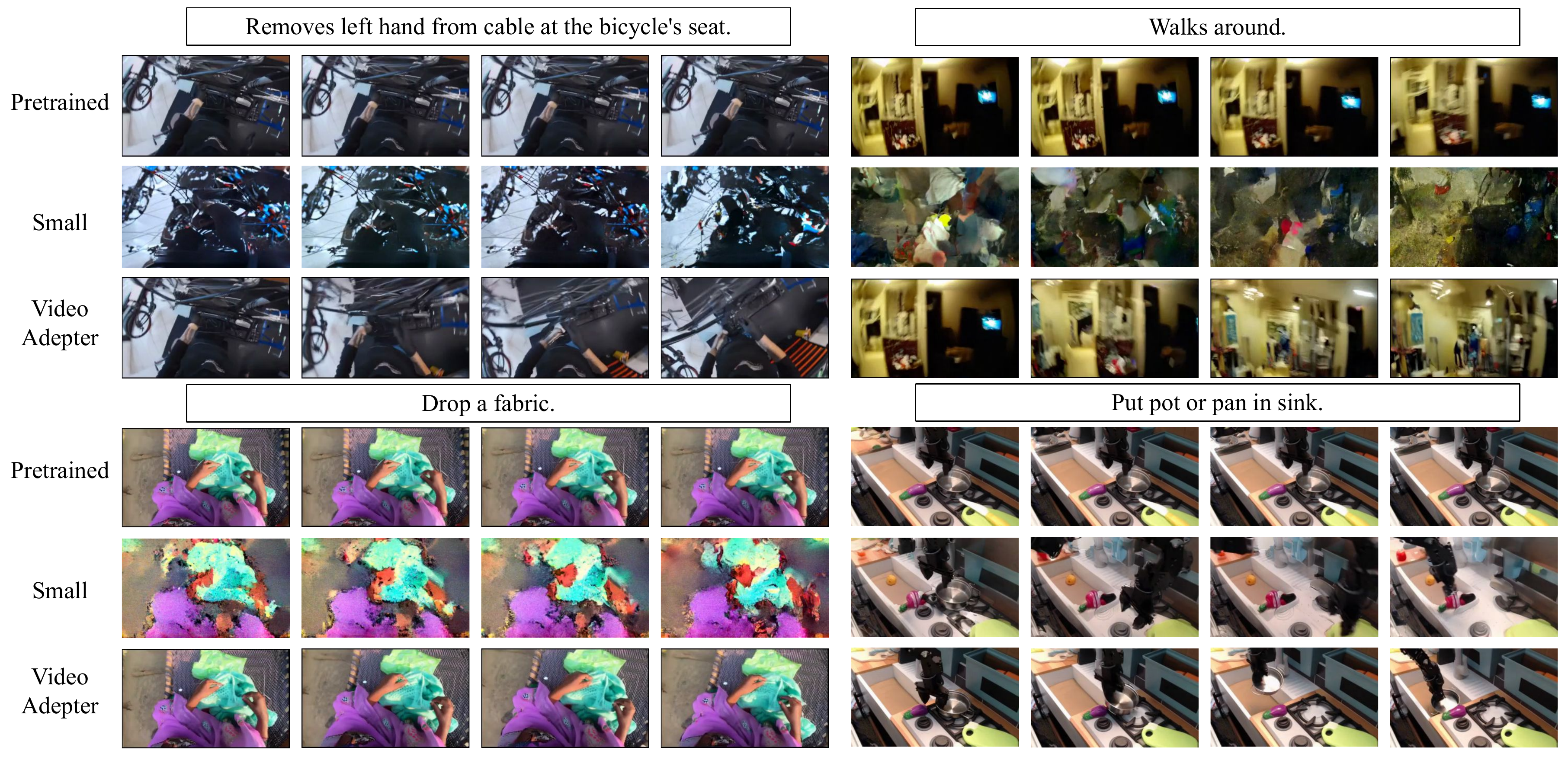}
\caption{\textbf{\methodname on Ego4D.} The pretrained model (first row) produces high-quality but nearly static videos that do not reflect the egocentric nature.
The small model (second row) produces low-quality videos but with more egocentric movements. \methodname (third row) produces high-quality and egocentric videos.}
\label{fig:video_ego4d}
\end{figure}

\myparagraph{Quantitative Results.} Table~\ref{tbl:video} shows the quantitative performance of \methodname under different video modeling metrics. On the Bridge Data, training a small model with parameters equivalent to 1.25\% of the pretrained video model (first row) already achieves better metrics than the pretrained model. However, \methodname incorporating the pretrained model as a probablistic prior is able to further improve the metrics of the small model (second row). On Ego4D, due to the complexity of the egocentric videos, the smallest model with 1.25\% of the pretrained video model can no longer achieve performance better than the pretrained model (first row), but incorporating the pretrained model during sampling still improves performance (second row). After increasing the size of the small model to half of the pretrained model, \methodname is able to achive better metrics than both the pretrained and task-specific model.

\myparagraph{Qualitative Results.} 
Figure~\ref{fig:video_bridge} and Figure~\ref{fig:video_ego4d} show the generated videos on Bridge Data and Ego4D. On the Bridge Data in Figure~\ref{fig:video_bridge}, the pretrained model produces videos that do not correspond to the task described by the text (there is no robot arm in the generated video). The task-specific small model produces videos with unrealistic movements that teleport the robot arm. \methodname, on the other hand, produces videos with realistic movements that successfully complete the task. On Ego4D in Figure~\ref{fig:video_ego4d}, the pretrained model produces high quality videos that contain little egocentric movement (first row), as the pretraining data mostly consists of generic videos from the internet that are not egocentric. The task-specific small model trained on Ego4D, on the other hand, produces videos with egocentric movement but of low quality (second row) due to limited model capacity. \methodname combines the best of both and generates high-quality egocentric videos (third row).

\begin{wrapfigure}{r}{5.0cm}
\small\setlength{\tabcolsep}{5.5pt}
\centering
\begin{tabular}{lcc}
      {\bf Model} & FVD $\downarrow$ & FID $\downarrow$ \\
      \midrule
      CFG Mix & 167.4 & 33.1 \\
      Small (L) & 152.5 & 30.1 \\
      \methodname & \textbf{148.1} & \textbf{29.5}\\
    \bottomrule
\end{tabular}
\caption{\small \textbf{Ablations.} \methodname improves the underlying video modeling performance of models on Bridge while directly mixing classifier-free scores  (CFG Mix)  hurts performance.}
\label{tbl:ablations}
\vspace{-20pt}
\end{wrapfigure}

\myparagraph{Ablations.} In \tbl{tbl:ablations}, we report generative modeling performance of the small model on Bridge either using \methodname, or a interpolation between the classifier-free scores of pretrained and small models. We find that \methodname improves performance, while interpolation between classifier-free scores hurts performance.

\subsection{Sim-to-Real Video Augmentation}

\begin{figure}[t]
\includegraphics[width=\textwidth]{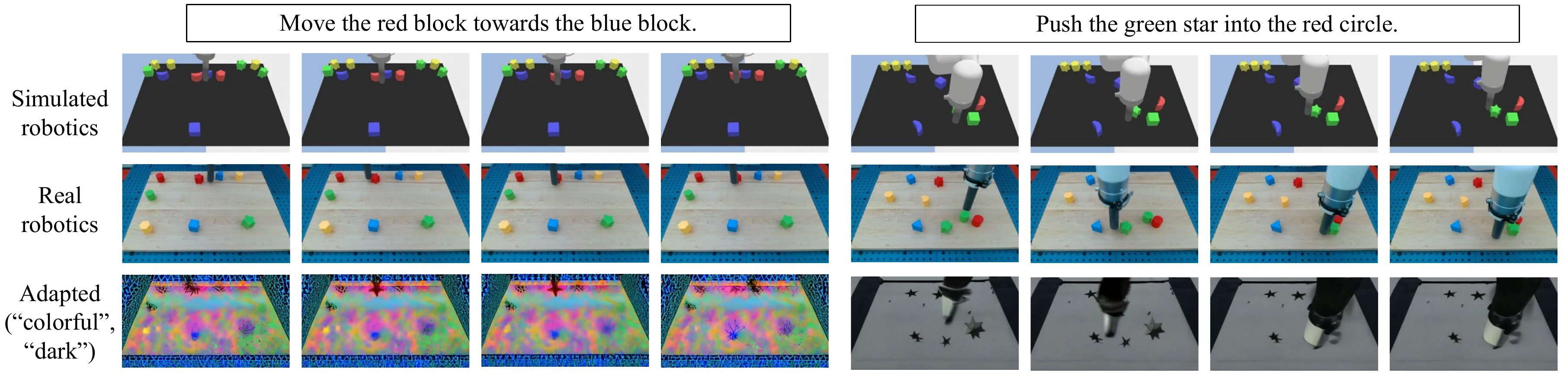}
\caption{\textbf{\methodname on sim-to-real transfer.} First row: simulated videos of execution plans generated by \methodname. Second row: real videos of execution plans generated by \methodname. Third row: real videos of execution plans generated by \methodname with data augmentation.}
\label{fig:video_sim2real}
\end{figure}

\myparagraph{Setup.} One important application of controllable video synthesis is to render realistic robotic videos from simulation with a variety of data augmentations so that policies trained on the augmented observations are more likely to generalize well to real-world settings~\citep{zhao2020sim}. To demonstrate \methodname's capability in sim-to-real transfer, we train a task-specific small edge-to-real model on 160k real robot trajectories of the LanguageTable dataset~\citep{lynch2022interactive},  generating videos of execution conditioned on the Sobel edges of the real videos. Similarly, we train another small edge-to-sim model on 160k simulated robot videos. Note that the simulated and real robotics data are not paired (paired sim-to-real data are hard to find) but are connected through edge-conditioning. We again leverage the edge-conditioned pretrained large model for customized stylisation.

\myparagraph{Adapted Videos.} Figure~\ref{fig:video_sim2real} shows the generated robotic videos from \methodname. \methodname can effectively generate paired simulated and real robotic videos that complete a task described by a language prompt, and further generate videos with various data augmentation styles that can be utilized to train policies with better sim-to-real transfer abilities through techniques similar to domain randomization~\citep{tobin2017domain}.
\section{Related Work}\label{sec:related}

\myparagraph{Text-to-Video Synthesis.} Following the recent success of text-to-image models~\citep{nichol2021glide,balaji2022ediffi,ramesh2022hierarchical,rombach2022high,saharia2022photorealistic,yu2022scaling,chang2023muse}, large text-to-video models with autoregressive~\citep{hong2022cogvideo,villegas2022phenaki,wu2022nuwa,wu2021godiva} and diffusion~\citep{ho2022imagen,singer2022make,blattmann2023align,zhou2022magicvideo,esser2023structure} structures have been developed, often by extending existing text-to-image models. Since video generation can be much more complex than image generation due to the additional temporal dimension, text-to-video models are often significantly larger in size. For instance, Imagen Video~\citep{ho2022imagen} requires a total of 11.6B parameters, and CogVideo~\citep{hong2022cogvideo} has 9B parameters. While various works have adopted latent diffusion to lower the computational overhead of video modeling~\citep{blattmann2023align,he2022latent,zhou2022magicvideo,esser2023structure}, these models still have a few billion parameters~\citep{blattmann2023align}, which poses a significant computational challenge to finetuning and domain specific adaptation.

\myparagraph{Adapting Pretrained Models} Adapting pretrained models for customized editing, inpainting, and stylization has been extensively studied in text-to-image and image-to-image translation models~\citep{gal2022image,hertz2022prompt,kawar2022denoising,li2022srdiff,lugmayr2022repaint,meng2021sdedit,ruiz2022dreambooth,saharia2022palette,sasaki2021unit,su2022dual}. In text-to-video models, most existing work on controlling generation behavior has been to either leverage text prompts~\citep{molad2023dreamix,esser2023structure}, finetuning a pretrained video model on stylized data~\citep{wu2022tune}, or performing light training on a copy of the pretrained video model similar to ControlNet~\citep{dhesikan2023sketching}. Solely relying on text description to specify video style can be unreliable when the desired style was not a part of any text description in the pretraining data. Finetuning pretrained video models, on the other hand, often gets too expensive as models get more complex. The idea of prompting as a training-free form of adaptation has been widely exploited in language models~\citep{liu2023pre}. Our work shares a similar spirit to language model prompting in that we do not finetune the weights of the pretrained video model. On the other hand, we use the pretrained model as a probablistic prior to achieve training free adaptation. Another set of work in adapting language models have trained a small controllable component similar to prefix-tuning~\citep{li2021prefix} and LoRA~\citep{hu2021lora}. \methodname's approach falls in the same category as these work but is designed specifically for text-to-video generation.

\myparagraph{Compositional Generative Models.} The techniques in this paper are further related to existing work on compositional generative modeling ~\citep{liu2022compositional,lace,du2020compositional,du2023reduce,du2021comet, liu2021learning,  wang2023concept,wu2022zeroc,deng2020residual,urain2021composable,gkanatsios2023energy,gandikota2023erasing,po2023compositional}, where different generative models are probabilistically combined to jointly generate outputs. In ~\citep{du2020compositional}, an approach to combine different probability distributions using EBMs is introduced. Most similar in spirit to this work, ~\citep{deng2020residual} composes a pretrained language model with a small EBM to improve language generation. However, different from this work, the small EBM is used to improve to global consistency of the language model, whereas we aim to use a small model to probabilistically adapt to a large pretrained video model to separate domains.
\section{Conclusion}
As  text-to-video foundation models grow even larger, effective adaptation of such models to task-specific usage is inevitable. We have introduced \methodname, an effective approach for using large pretrained text-to-video models as a probabilistic prior that can guide generation of domain and task specific videos. \methodname does not require any finetuning of the large pretrained model, yet is able to utilize knowledge from the pretrained model to synthesize higher quality videos in specific domains or with particular styles of interest. 

\paragraph{Limitations and Broader Impact.}
While \methodname is proposed as a way to efficiently adapt large pretrained text-to-video models, it is not completely training free, as \methodname still requires a small video model to be trained on domain-specific data. Additionally, \methodname requires outputting the score in addition to the generated video, which is not standard for existing text-to-image and text-to-video APIs. Since the lack of open access to model weights and computational efficiency are the main motivations of \methodname, \methodname effectively makes text-to-video research more accessible to small industrial and academic institutions.

\begin{ack}
    Thanks to Wilson Yan, Oleg Rybkin, Hanjun Dai, and Douglas Eck for reviewing draft versions of this manuscript. We gratefully acknowledges the support of a Canada CIFAR AI Chair, NSERC and Amii, NSF GRFP, and support from Berkeley BAIR industrial consortion. 
\end{ack}

\bibliography{neurips_2023}
\bibliographystyle{unsrt}

\appendix
\clearpage
\begin{center}
{\huge Appendix}
\end{center}

In the Appendix we provided a detail derivation of connection between diffusion models and EBMs in \sect{app:diffusion-ebm}. We further provide additional experimental details in \sect{app:exp}. Finally, we provide a comparison with using the same computational budget to finetune the existing large pretrained model in \sect{sect:finetune}.

\section{Connection between Diffusion and EBM}
\label{app:diffusion-ebm}
The sampling procedure in a diffusion model corresponds to the Langevin sampling procedure on an EBM. To see this, we consider perturbing a sample $\tau^{t-1}\sim p\rbr{\tau^{t-1}}$ from target distribution $p(\tau^{t-1})$ with a Gaussian noise, \ie, 
\[
\tau^{t} = \tau^{t-1} + \xi, \quad \xi\sim \Ncal\rbr{\textbf{0}, \sigma_t^2\Ib}
\]
which corresponds to the transition operator 
\[
\Tcal(\tau^{t}|\tau^{t-1}) \propto \exp\rbr{\frac{-\nbr{\tau^t - \tau^{t-1}}^2}{2\sigma_t^2}}
\]
where the joint distribution of $\tau^t$ and $\tau^{t-1}$ is
\[
p(\tau^t, \tau^{t-1}) \propto \exp\rbr{\psi\rbr{\tau^{t-1}} - \frac{\nbr{\tau^t - \tau^{t-1}}^2}{2\sigma_t^2}}.
\]

We can express the Bayes estimator of $\tau^{t-1}$ given the perturbed observation $\tau^t$ as
\begin{align}\label{eq:bayes_mean}
    m(\tau^t) &= \int \tau^{t-1} p_\theta(\tau^{t-1}|\tau^t)d\tau^{t-1}  = \tau^t + \sigma_t^2\nabla \log p\rbr{\tau^t} 
\end{align}
\begin{proof}
By the property of Gaussian distribution, we have
    \begin{eqnarray}
    \sigma^2\nabla_{x'}p\rbr{x'|x} = p\rbr{x'|x} \rbr{x - x'}.
    \end{eqnarray}
    Therefore, we have
    \begin{eqnarray}
        &&\sigma \nabla_{x'}\int p\rbr{x'|x}p(x)dx = \int \rbr{x - x'}p\rbr{x', x} dx = \int xp\rbr{x', x}dx  - x'p(x')\\
        &\Rightarrow & \sigma \nabla_{x'}\log p\rbr{x'} = \int x\frac{p\rbr{x' , x}}{p\rbr{x'}} dx - x' = \EE\sbr{X|x'} - x'
    \end{eqnarray}
\end{proof}

Thus, we can represent the perturbed data with an EBM $p(\tau^t)\propto \exp\rbr{E_\theta\rbr{\tau^t, \sigma_t}}$, and learn the parameters through regression ~\citep{vincent2011connection,saremi2018deep,saremi2019neural,song2019generative}, which leads to the optimal solution
\begin{align}\label{eq:mean_regression}
    &\min_\theta\,\, \EE_{\tau^{t-1}\sim\Dcal, \xi\sim \Ncal\rbr{0, \sigma_t^2\Ib}}\sbr{\nbr{\tau^{t-1} - m(\tau^{t})}^2} \nonumber \\&= \EE_{\tau^{t-1}\sim\Dcal, \xi\sim \Ncal\rbr{0, \sigma^2\Ib}}\sbr{\nbr{- \xi - \nabla E_\theta\rbr{\tau^{t-1} + \xi, \sigma_t})}^2},
\end{align}
whch also corresponds to the denoising diffusion training objective.

Once we have the trained $E_\theta\rbr{\tau^t}$, we can then recover the sample $\tau^{t-1}$ according the denoising sampling procedure
\begin{eqnarray}\label{eq:stoc_localization}
    \tau^{t-1} = \alpha^t m(\tau^t) + \alpha^t\xi = \alpha^t(\tau^t - \gamma\nabla_{\tau^t} E_{\theta}\rbr{\tau^t,\sigma_t} ) + \alpha^t \xi, \quad \xi \sim \Ncal\rbr{\textbf{0}, \sigma_t^2\Ib}
\end{eqnarray}
which corresponds to the sampling via stochastic localization~\citep{el2022sampling} and Equation~\ref{eq:unconditional_ebm} in the main paper.

\section{Experimental Details}
\label{app:exp}
\subsection{Experiment Details}
\paragraph{Dataset} The large pretrained model is trained on 14 million video-text pairs plus 60 million image-text pairs, and with the LAION-400M image-text dataset. The images are spatially resized to 24x40 and videos using anti-aliased bi-linear resizing. We use different frame rate for different types of videos for best visualization results. For the Bridge~\citep{ebert2021bridge} we directly use the released opensource dataset. For Ego4D~\citep{grauman2022ego4d} data, we take a small portion of the released dataset. For Anime and Sci-Fi style, we curate two separates datasets with their respective keywords. The keywords used for filtering data for Anime style are (in small letter) ``disney'', ``cartoon'', ``anime'', ``animation'', ``comic'', ``pixar'', ``animated'', ``fantasy''. The keywords used for filtering data for Sci-Fi style are ``science fiction'', ``sci-fi'', ``scifi'', ``astronaut'', ``alien'', ``NASA'', ``interstellar''. For the animation with a particular artist style, we use the Case Closed animation (also named Detective Conan). For the Language Table dataset, we used the data from \citep{lynch2022interactive}.

\begin{table*}[h!]
\small\setlength{\tabcolsep}{4pt}
\centering
\begin{tabular}{lcccccccc}
      {\bf Dataset} & Pretrain & Bridge & Ego4D & Anime & Sci-Fi & Case Closed & LangTable Sim & LangTable Real \\
      \midrule
      \# Train & 474M & 2.3k & 97k & 0.6M & 21k & 5k & 0.16M & 0.16M \\
    \bottomrule
\end{tabular}
\caption{\small \textbf{Training data size.} Number of text-video or text-image pairs used for training the pretrained large model and each of the small model. Training data for particular styles can be magnitude smaller than the pretraining dataset.}
\label{tbl:training_samples}
\end{table*}

\paragraph{Architecture.} To pretrain the large model, we use the same pretraining dataset, base architecture, and training setup as \citep{ho2022imagen}, with modifications of first-frame conditioning for Bridge and Ego4D, and edge conditioning for stylisation and sim-to-real. Specifically, the large model architecture consists of video U-Net with 3 residual blocks of 1024 base channels and channel multiplier [1, 2, 4], attention resolutions [6, 12, 24], attention head dimension 64, and conditioning embedding dimension 1024. To support first frame conditioning, we replicate the first frame across all future frame indices, and concatenate the replicated first frame channel-wise to the noisy data following ~\citep{du2302learning}. To support edge conditioning, we run a sobel edge detector and use gradient approximations in the x-direction as the conditional video, and concatenates these edge frames with noisy data similar to first-frame conditioning. The large model consists of 5.6 billion parameters in total. For the set of small models for adaptation, Ego4D Small (L) has 512 base channels in each of the residual blocks. Ego4D Small (S) and Bridge Small (S) have a single residual block with 32 base channels. Bridge Small (L) has a single residual block with 64 base channels. The set of stylisation models (animation, sci-fi, and particular anime style) have 3 residual blocks and 256 base channels. For illustrating the generated videos at a higher resolution, we train two additional spatial super resolution models 24x40 $\rightarrow$ 48x80 (1.4B) and 48x80 $\rightarrow$ 192x320 (1.2B). We additionally use T5-XXL~\citep{raffel2020exploring} to process input text prompts which consists of 4.6 billion parameters, which we omit from the parameter count as all large and small models require text embeddings.

\paragraph{Training and Evaluation.} We train each of our video diffusion models for 2M steps
using batch size 2048 with learning rate 1e-4 and 10k linear warmup steps. The large 5.6B pretrained model requires 512 TPU-v4 chips, whereas various small models require anywhere between 8 and 256 TPU-v4 chips depending on the size. We use noise schedule log SNR with range [-20, 20]. We use 128 samples and 1024 samples to compute the FVD, FID, and Inception Scores metric on Bridge and Ego4dD, respectively.

\paragraph{Sampling.}  All diffusion models are trained with 1000 timesteps of sampling. To generate videos, we combined scores from both pretrained models and adapter models for all timesteps except the last 100 timesteps. The last 100 timesteps capture high frequency information in an image, and we found better image quality if we did not combine scores in these timesteps. We use a pretrained neural strength of 0.2 for Ego4D and 0.1 for Bridge, and 0.4 for all animation datasets.

\clearpage
\newpage

\section{Comparison to Finetuning}
\label{sect:finetune}

To illustrate the computational efficiency of \methodname, we further compare video modeling metrics of \methodname to finetuning the pretrained model for an equivalent number of TPU time. Specifically, the pretrained model requires 512 TPU-v4 chips whereas the small model on Bridge data requires 8 TPU-v4 chips. The small Bridge model requires 100k steps to reach convergence, and hence we finetune the pretrained model for 100,000 / 64 = 1,560 steps. The finetuning results are shown in Table~\ref{tbl:app_video}. \methodname achieves better FVD and FID than finetuning the pretrained model for an equal number of TPU steps.

\begin{table*}[h]
\small\setlength{\tabcolsep}{5.5pt}
\centering
\begin{tabular}{lccc}
      &  \multicolumn{3}{c}{\bf Bridge} \\
      \cmidrule(lr){2-4}
      {\bf Model} & FVD $\downarrow$ & FID $\downarrow$ & Param (B)$\downarrow$ \\
      \midrule
      Small (S) & 186.8 & 38.8 & 0.07 \\
      Small (S) + Pretrained & \textbf{177.4} & \textbf{37.6} & 0.07\\
      Small (L) & 152.5 & 30.1 & 0.14 \\
      Small (L) + Pretrained & \textbf{148.1} & \textbf{29.5} & 0.14\\
      Pretrained & 350.1 & 42.6 & 5.6\\ 
      Pretrained Finetune & 321.0 & 39.4 & 5.6 \\
    \bottomrule
\end{tabular}
\caption{\small \textbf{Video Modeling Quantitative Performance} \methodname achieves better FVD and FID than finetuning the pretrained model for equal number of TPU steps.}
\label{tbl:app_video}
\end{table*}

\end{document}